\documentclass[a4paper]{article}

\usepackage{times}
\normalfont
\usepackage[english]{babel}
\usepackage{graphicx}
\usepackage{amsmath,amssymb,stmaryrd,mathtools}
\usepackage[utf8]{inputenc} 
\usepackage{algorithmic,algorithm}
\usepackage{xspace}

\newcommand{\tes}{$TES_\theta$\xspace}

\begin{document}

\title{Analysis of attractor distances\\ in Random Boolean Networks}

\author{Andrea Roli and Stefano Benedettini\\DEIS--Cesena\\ \textit{Alma Mater Studiorum} Universit\`{a} di Bologna\\ Italy \and Roberto Serra and Marco Villani\\Dipartimento di scienze sociali, cognitive e quantitative\\ Universit\`a di Modena e Reggio Emilia, Italy\\ and\\ European Centre for Living Technology, Venice, Italy}

\date{}

\maketitle

\begin{abstract}
We study the properties of the distance between attractors in Random Boolean Networks, a prominent model of genetic regulatory networks. We define three distance measures, upon which attractor distance matrices are constructed and their main statistic parameters are computed. The experimental analysis shows that ordered networks have a very clustered set of attractors, while chaotic networks' attractors are scattered; critical networks show, instead, a pattern with characteristics of both ordered and chaotic networks.
\end{abstract}

\section{Introduction}

Boolean networks (BNs) have been introduced as models of genetic regulatory networks (GRNs) by Kauffman~\cite{Kauf93}. Their interest as GRN models relies primarily in the fact that some classes of BNs statistically reproduce some characteristics of real cells. For example, it has been shown that single gene knock-out experiments can be simulated in Random BNs~\cite{SerVilSem-theorbio}. Reproducing statistical properties of real cells through specific classes of GRNs is called the {\em ensemble approach}, that was proposed by Kauffman~\cite{Kauf-ensemble}. In this approach, the objective of the modelling process is not to define a model for a single, specific cell, but rather to reproduce the statistics of the parameters of interest of the ensemble of cells of the given type.
In general, for an ensemble of real cells it is possible to define a set of features, such as some parameter on cell dynamics in case of perturbation. An ensemble of GRNs is of interest if the matching between its features and those of real cells is high.
Nowadays, modern biotechnology tools, such as DNA microarrays, make it possible to gather a huge amount of biological data, hence the ensemble approach can be applied even more effectively than in the past.

A prominent feature that can be considered in the ensemble approach is the distribution in distances between gene expression levels in different types of cells. It was conjectures by Kauffman~\cite{Kauf93} that attractors in Random BNs correspond to cellular types, a conjecture further refined in terms of {\em threshold ergodic sets} by Serra et al. in~\cite{Serra-TES}. In order to test this conjecture, two issues have to be addressed: first, the properties of BN attractors have to be studied; second, these properties have to be compared with the ones of cellular types.
In this work, we aim at providing a contribution to the first issue by studying the statistics of distances between attractors in Random BNs. In Section~\ref{sec:bn} we briefly summarise the main concepts and results in the field of BNs with emphasis on Random BNs. We then discuss measures and features of interest for the attractors in BNs in Section~\ref{sec:attractors}. Results of an experimental analysis are presented and discussed in Section~\ref{sec:experiments}. We finally summarise this contribution and outline future research in Section~\ref{sec:conclusion}.

\section{Boolean networks}
\label{sec:bn}

BNs have been firstly introduced by Kauffman~\cite{Kauf93} and subsequently received considerable attention in the composite community of complex systems research. Recent advances in this field can be mainly found in works addressing themes in genetic regulatory networks or investigating properties of BNs themselves~\cite{aldana-robust,fretter-response,ribeiro-mutual-information,SerVilGraKau-theorbio}.

A BN is a discrete-state and discrete-time dynamical system defined by a directed graph of $n$ nodes, each associated to a Boolean variable $x_i$, $i = 1,\ldots,n$, and a Boolean function $f_i(x_{i_1},\ldots,x_{i_{k_i}}$), where $k_i$ is the number of inputs of node $i$. Often, $k_i$ is chosen to be equal to a constant value $k$ for every $i$. The arguments of the Boolean function $f_i$ are the nodes whose outgoing arcs are connected to node $i$. The state of the system at time $t$, $t \in \mathbb{N}$, is defined by the array of the $n$ Boolean variable values at time $t$: $s(t) = \langle x_1(t), \ldots, x_n(t) \rangle$. The most studied dynamics for BNs is \textit{synchronous}, i.e., nodes update their states in parallel, and \textit{deterministic}. However, many variants exists, including asynchronous and probabilistic update rules~\cite{Gershenson2004c}.

In this work, we consider networks ruled by synchronous and deterministic dynamics. Given this setting, the network trajectory in the $n$-dimensional state space is a sequence of states composing a \textit{transient}, possibly empty, followed by an \textit{attractor}, that is a cycle of length $l \in [1,\ldots, 2^n]$.
When BNs are employed as genetic regulatory network models, attractors assume a notable relevance as they can be interpreted as cellular types~\cite{huang-breast-desease2007}. This interpretation has recently been extended by considering sets of attractors, the so-called {\em Threshold Ergodic Sets} (\tes), instead of single attractors~\cite{Serra-TES}. This extension provides support to the usefulness of RBNs as GRN models, as it makes it possible also to model cell differentiation dynamics.

A special category of BNs that has received particular attention is that of Random BNs, which can capture relevant phenomena in genetic and cellular mechanisms and complex systems in general. Random BNs (RBNs) are usually generated by choosing at random $k$ inputs per node and by defining the Boolean functions by assigning to each entry of the truth tables a 1 with probability $p$ and a 0 with probability $1-p$. Parameter $p$ is called \textit{homogeneity} or \textit{bias}. Depending on the values of $k$ and $p$ the dynamics of RBNs is \textit{ordered} or \textit{chaotic}. In the first case, the majority of nodes in the attractor is frozen and any moderate-size perturbation is rapidly dampened and the network returns to its original attractor. Conversely, in chaotic dynamics, attractor cycles are very long and the system is extremely sensitive to small perturbations: slightly different initial states lead to divergent trajectories in the state space. RBNs temporal evolution undergo a second order phase transition between order and chaos, governed by the following relation between $k$ and $p$: $k_c = [2 p_c (1 - p_c)]^{-1}$, where the subscript $c$ denotes the critical values~\cite{derrida1986}. Networks along the \textit{critical} line have important properties, such as the capability of achieving the best balance between evolvability and robustness~\cite{aldana-robust} and maximising the average mutual information among nodes~\cite{ribeiro-mutual-information}.

\section{Attractor distance statistics}
\label{sec:attractors}

In real cells, each type is characterised by a specific pattern of gene expression levels which can be represented as real number vectors of size $n$, where $n$ is the number of genes.
On the model side, we can make the hypothesis that each attractor of a BN represents a cell type.
Statistics and, possibly, other kinds of information on the distances between attractors can be computed and then compared against equivalent statistics on gene expression levels in real cell types, so as to test to what extent the class of RBNs capture relevant properties of ensembles of real cells.\footnote{The hypothesis of the correspondence between attractors and cell types is therefore operational, rather than ontological.}

In this Section, we introduce the distance measures we defined over the attractors, along with the statistics and properties we analysed.

\subsection{Attractor distance measures}

We defined and studied three different distances among BN attractors, namely {\em min-Hamming}, {\em Euclidean} and {\em pseudo-Hamming}. The first one is defined upon the states composing the attractors, while the other two are defined upon the average values of BN nodes in each attractor.

\vspace{2ex}

\noindent
{\bf min-Hamming}. This distance measures the minimum number of node values that should be changed in order to let the network's trajectory jump from an attractor directly to another one. Formally:
\begin{displaymath}
d_{mH}({\cal A}_i,{\cal A}_j) = {\rm min}\{H_d(s,s') : s \in {\cal A}_i, s' \in {\cal A}_j)\}
\end{displaymath}
where $H_d(s,s')$ is the Hamming distance between states $s$ and $s'$.
It is important to observe that this distance does not depend on the network dynamics in the state space, as it simply considers the Hamming distance between states independently of the state space trajectory. Measures which depend on the actual state space topology can be also defined.

\vspace{2ex}

\noindent
The following distances are defined over real vectors $V({\cal A}_i) = \langle v_1,\ldots,v_n \rangle$, each one computed for a given attractor ${\cal A}_i$. Elements $v_j$ ($j = 1,\ldots, n$) are computed by averaging the values assumed by variable $x_j$ along the attractor, i.e., by computing the fraction of times a Boolean variable assumes value 1 along the attractor. In formulas: given attractor ${\cal A} = (s^{(1)},\dots,s^{(\tau)})$ of period $\tau$, with $s^{(h)} = \langle x^{(h)}_1,\ldots,x^{(h)}_n \rangle , h = 1,\ldots,\tau$, each element $v_j$ of vector $V({\cal A})$ is computed as: 
$v_j = \frac{1}{\tau} \sum_{h = 1}^{\tau}{x^{(h)}_j}$.
It has to be noted that this mapping between attractors and vectors of real numbers makes it possible to establish a simple yet direct semantics of a BN attractor as a gene expression level array~\cite{SerVilGraKau-theorbio}.

\vspace{1ex}

\noindent
{\bf Euclidean}. A straightforward way of measuring the distance between two real valued vectors is to compute their Euclidean distance. This distance induces naturally a distance over attractors:
\begin{displaymath}
d_{Eucl}({\cal A}_i,{\cal A}_j) = d_{Eucl}(V_i,V_j) = \sqrt{\sum_{l = 1}^n{(v_{il} - v_{jl})^2}}
\end{displaymath}

\vspace{1ex}

\noindent
{\bf pseudo-Hamming}. The Euclidean distance might smooth the differences between expression vectors, thus making it hard to distinguish between attractors of different length. In fact, attractor cycles of very different length might be mapped onto real valued vectors whose Euclidean distance is very small.
For this reason, we introduced a distance that is computed by summing up the number of homologous vector entries which are different. In formulas:
\begin{displaymath}
d_{{\psi}H}({\cal A}_i,{\cal A}_j) = d_{{\psi}H}(V_i,V_j) = \sum_{l = 1}^n{1 - \delta(v_{il} - v_{jl})} \; \textrm{, where} \;
\delta(x,y) = \left\{ \begin{array}{ll}
 1 & \textrm{, if $x = y$}\\
 0 & \textrm{, otherwise}\\
  \end{array} \right.
\end{displaymath}

\subsection{Attractors clustering}

Given the attractors of a BN network, a \emph{distance matrix} can be constructed according to the distances previously defined. Besides computing the main statistical parameters of such data, distance matrices have been also used in two kinds of analysis: ({\em i}) distribution in (weighted) clustering coefficient and ({\em ii}) attractor dendrograms.\footnote{Preliminary results have been published in~\cite{roli-wivace2009}.}

\subsubsection{Clustering Coefficient}

The clustering coefficient $C_i$ of a node $i$ in a graph provides an estimation of the how much its neighbours tend to form a complete graph. For a non-weighted graph, the clustering coefficient $C_i$ is equal to its maximum value 1 if neighbours of $i$ form a complete graph, while it is 0 if neighbours of $i$ are disconnected. The average of node clustering coefficient provides an estimation of how much a graph is characterised by clusters of nodes. Formally, a network clustering coefficient is:
\begin{displaymath}
C = \frac{1}{N} \sum_{i = 1}^{N} C_i \qquad C_i = \frac{n_i}{g_i}
\end{displaymath}
where $n_i$ is the number of edges between neighbours of node $i$ and $g_i$ the maximum possible number of edges between them. It is also possible to extend the clustering coefficient definition to weighted graphs~\cite{Zhang-clustering}; in this case, the greater the edge weight, the stronger is the intensity of the connection between the two nodes. Values used for computing this measure are taken from a network adjacency matrix $A = (a_{ij})$, where an element $a_{ij}$ corresponds to the weight of the edge which has its tail in $i$ and its head in $j$; $a_{ij} = 0$ if $i = j$ or edge $(i,j)$ is not present. In formulas:
\begin{eqnarray*}
  n_i = \frac{1}{2} \sum_{u \neq i} \sum_{\{v \, | \, v \ne i, v \neq u\}} a_{iu} a_{uv} a_{vi} \qquad \textrm{,} \qquad
  g_i = \frac{1}{2} ((\sum_{u \neq i} a_{iu})^2 - \sum_{u \neq i} a^2_{iu})
\end{eqnarray*}
In our analysis, the weight of an edge that connects two nodes (attractors) is the (normalised) reciprocal of the distance between the attractors.

\subsubsection{Dendrograms}

A network attractor distance matrix can be also used to graphically represent clustered distribution of attractors. For each network, a dendrogram has been generated, which represents in a single data structure all the possible clusters of the elements in a set. Attractor dendrogram analysis yields a graphical representation of the tendency of the attractors to gather into clusters. The results we present are based on dendrograms constructed using `single-link' algorithm~\cite{Jain-dendrogram}.

\section{Experimental analysis}
\label{sec:experiments}

In this Section, we present the results of the experimental analysis performed by simulating BNs with `The Boolean Network Toolkit'~\cite{BNToolkit}.
We analysed the main statistical parameters of the distances between attractors in RBNs with 70 nodes,\footnote{Networks' size was constrained by the very large computational time required for simulating chaotic networks of larger size.} $k = 3$ and bias values such that the networks are in ordered ($p=0.85$), chaotic ($p=0.5$) and critical ($p=0.788675$) phases. 
For each parameter configuration, 50 independent RBN realisations have been generated. Each network dynamics has been simulated for at most $10^6$ steps, starting from $10^5$ initial states picked uniformly at random in order to sample attractor cycles.

\subsection{Distance measures statistics}

\begin{table}[t]
\caption{Distance statistics}\label{tab:distances}
\begin{center}
\begin{tabular}{|r||c|c|c|c|c|c|}
\hline
\multicolumn{7}{|c|}{\bf Minimum Hamming distance}\\
\hline
\textbf{Bias} & \textbf{Min.} & \textbf{1st Qu.}  & \textbf{Median} & \textbf{Mean} & \textbf{3rd Qu.} & \textbf{Max.} \\
\hline
0.5 & 1 &  5 &   9 &   9.03 & 12 &  29\\
0.788675 & 1 &   1 &   4 &   4.62 &  7 &  24\\
0.85 & 1 &   1 &   2 &   3.86 &  5 &  18\\
\hline
\end{tabular}\\
\vspace{1ex}
\begin{tabular}{|r||c|c|c|c|c|c|}
\hline
\multicolumn{7}{|c|}{\bf Euclidean distance between activation vectors}\\
\hline
\textbf{Bias} & \textbf{Min.} & \textbf{1st Qu.}  & \textbf{Median} & \textbf{Mean} & \textbf{3rd Qu.} & \textbf{Max.} \\
\hline
0.5 & 0.00 & 0.43 & 0.90 & 1.23 & 1.80 & 4.77\\
0.788675 & 0.00 & 0.36 & 1.23 & 1.18 & 1.74 & 4.69\\
0.85 & 0.00 & 0.67 & 1.16 & 1.36 & 1.88 & 4.24\\
\hline
\end{tabular}\\
\vspace{1ex}
\begin{tabular}{|r||c|c|c|c|c|c|}
\hline
\multicolumn{7}{|c|}{\bf Pseudo-Hamming distance between activation vectors}\\
\hline
\textbf{Bias} & \textbf{Min.} & \textbf{1st Qu.}  & \textbf{Median} & \textbf{Mean} & \textbf{3rd Qu.} & \textbf{Max.} \\
\hline
0.5 & 0 &   66 &   68 &   63.44  & 70 &   70\\
0.788675 & 0 &    3 &   12 &   13.61 &  22 &   51\\
0.85 & 0 &    3 &    8 &    8.35 &  10 &   27\\
\hline
\end{tabular}
\end{center}
\end{table}

A first analysis concerns the main statistical parameters of the distance matrices. Table~\ref{tab:distances} shows the minimum, maximum, mean, median and 1st and 3rd quartile values of such quantities.
We can observe that the maximal distances, independently of the actual definition used, are in chaotic networks. Moreover, also the mean and median values of attractor distance in chaotic networks are considerably higher than those of critical and ordered networks. The differences between the last two classes are smaller than those with respect to chaotic ones, even though critical networks show a larger spread in values and higher average values.\footnote{An exception to this observation is the median of the Euclidean distance, but differences are very small and not significant.}
It is remarkable to observe that the qualitative pattern is the same, independently of the distance measure.

\subsection{Attractors clustering}

\begin{figure}[t]
\begin{center}
\includegraphics[scale=0.3,angle=-90]{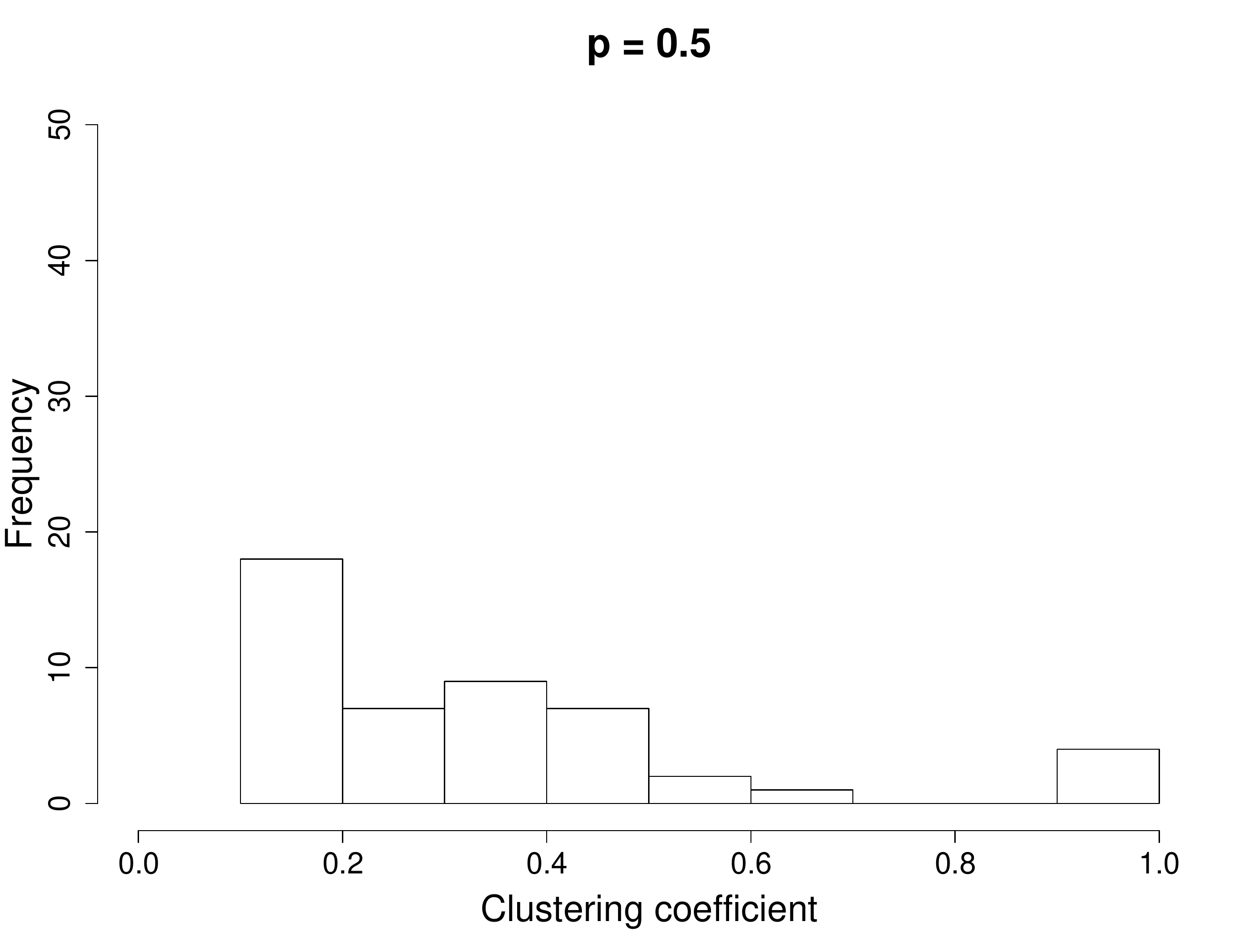}\\
{\small (a)}\\
\includegraphics[scale=0.3,angle=-90]{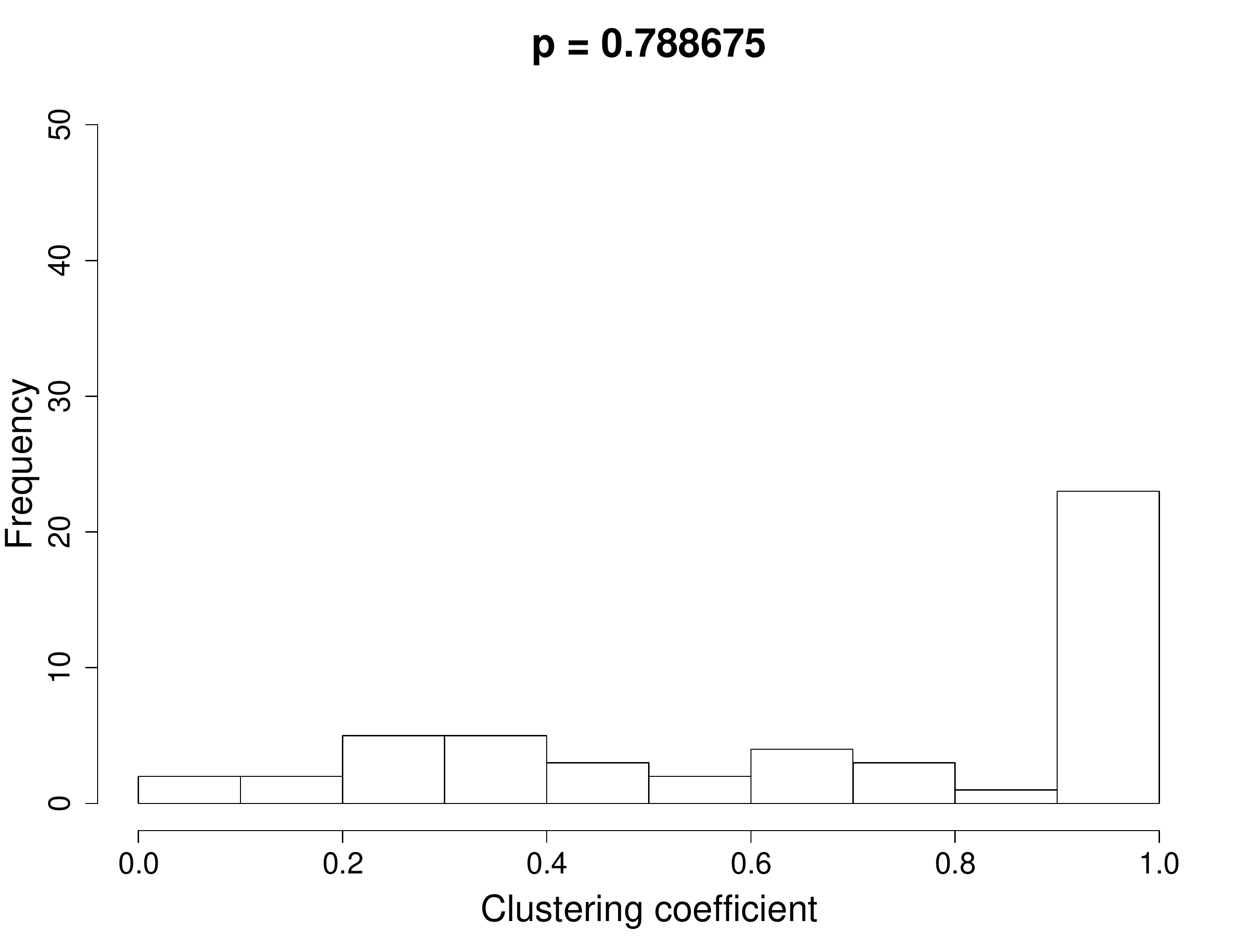}\\
{\small (b)}\\
\includegraphics[scale=0.3,angle=-90]{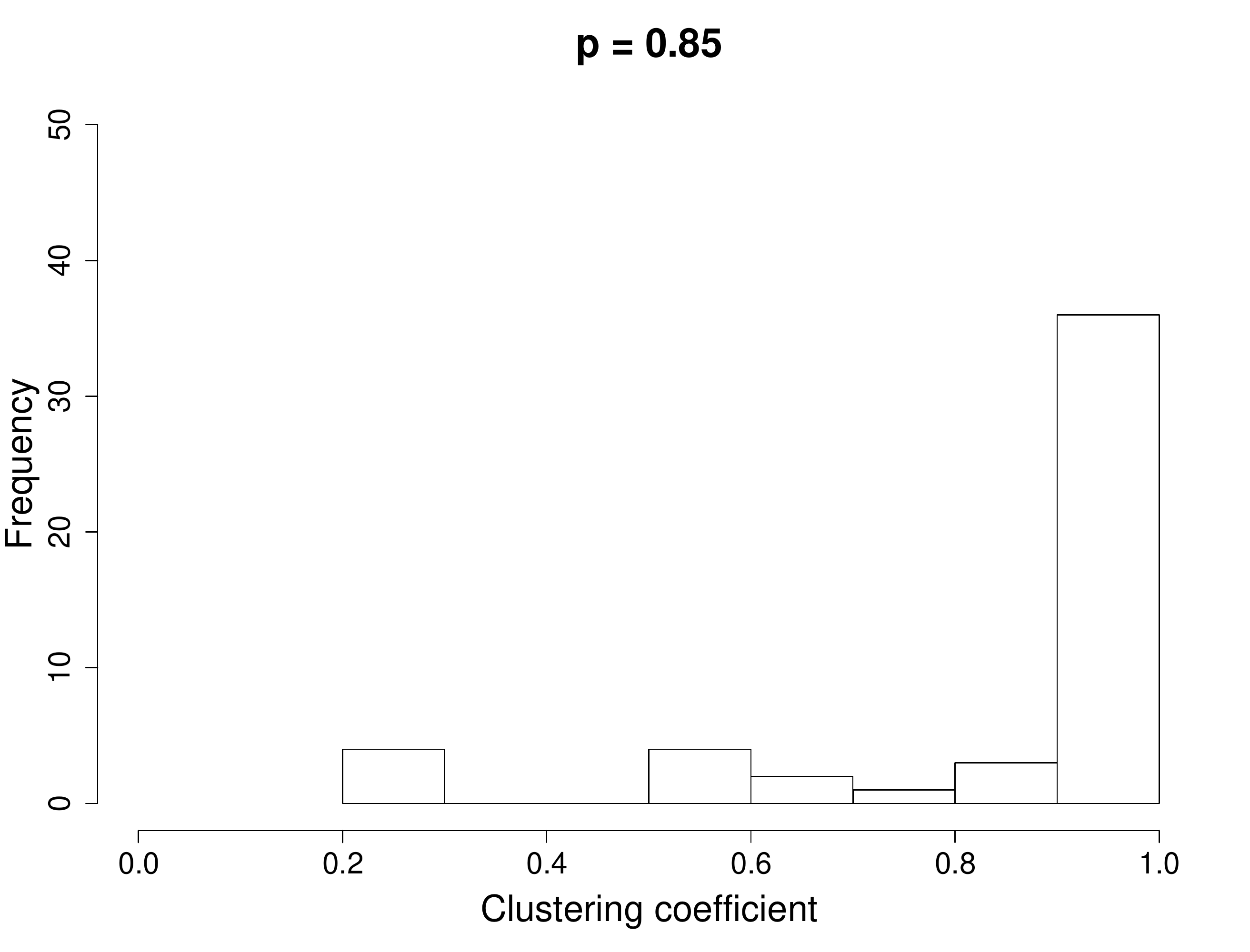}\\
{\small (c)}\\
\end{center}
\caption{Average clustering coefficient distribution}
\label{fig:clustering}
\end{figure}

In Figures~\ref{fig:clustering}(a), \ref{fig:clustering}(b) and \ref{fig:clustering}(c), the histogram of the average clustering coefficient distribution is plotted for chaotic, critical and ordered BNs, respectively. The distance measure considered is the {\em min-Hamming}, but qualitatively analogous results have obtained also with the other distance measures.
The pattern emerging from the histograms is not surprising: chaotic network attractors have a very low tendency of forming clusters, while in critical and ordered networks, attractors are clearly clustered. It is interesting to note that critical networks seem to exhibit a pattern that is a mixture of the chaotic and ordered ones, because the clustering coefficient distribution spans, with significant values, across the whole range.
\begin{figure}[t]
\begin{center}
\includegraphics[scale=0.3,angle=-90]{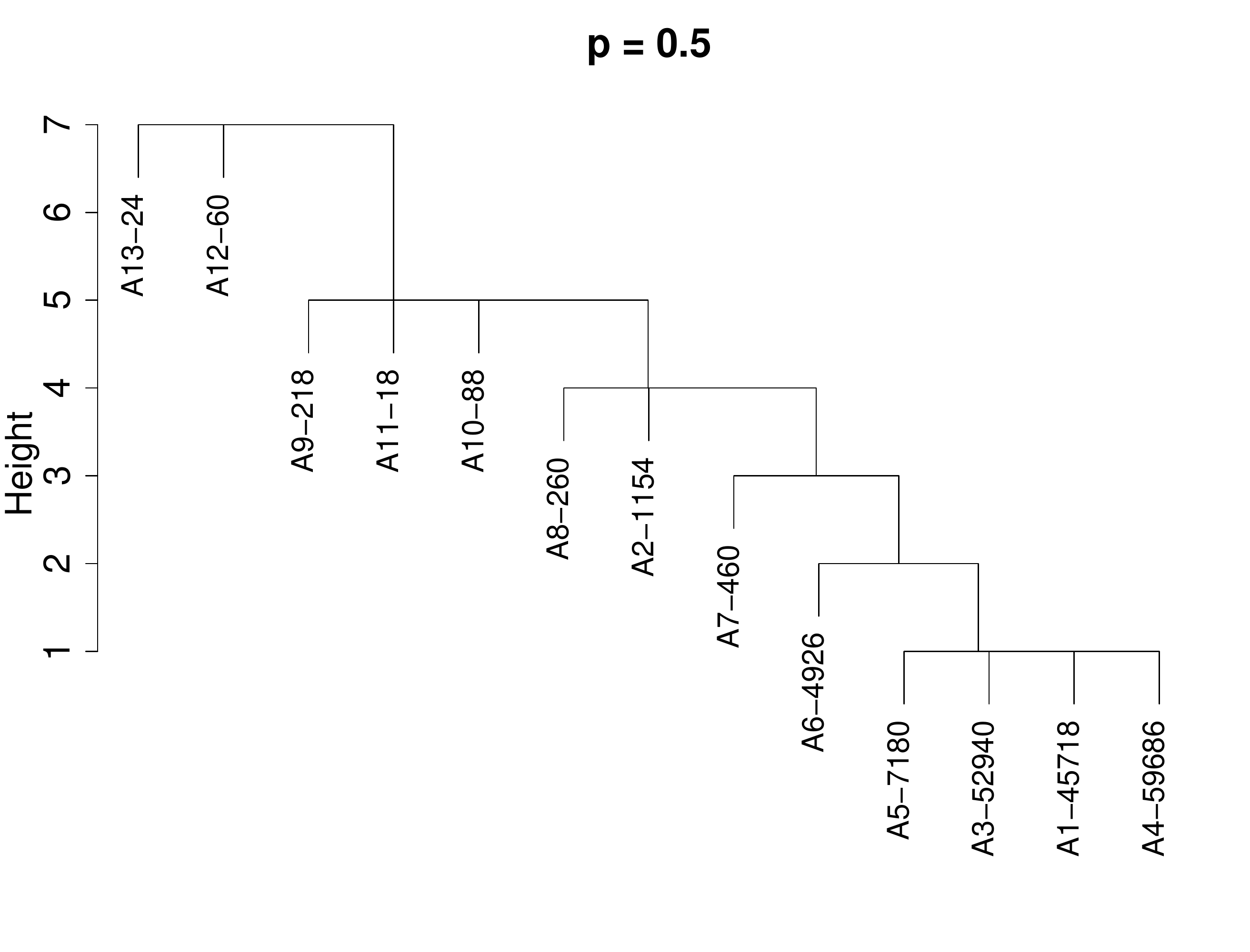}\\
{\small (a)}\\
\includegraphics[scale=0.3,angle=-90]{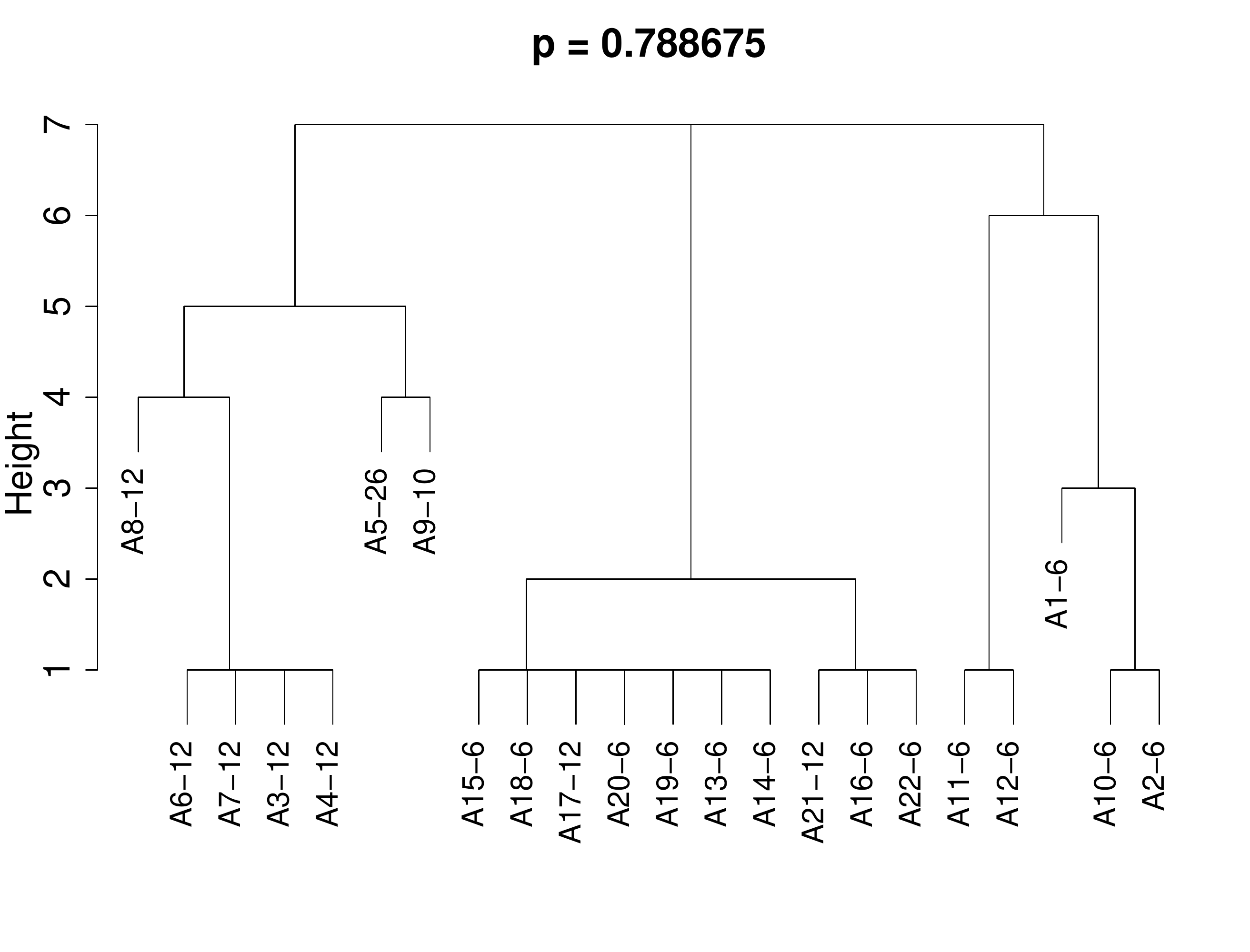}\\
{\small (b)}\\
\includegraphics[scale=0.3,angle=-90]{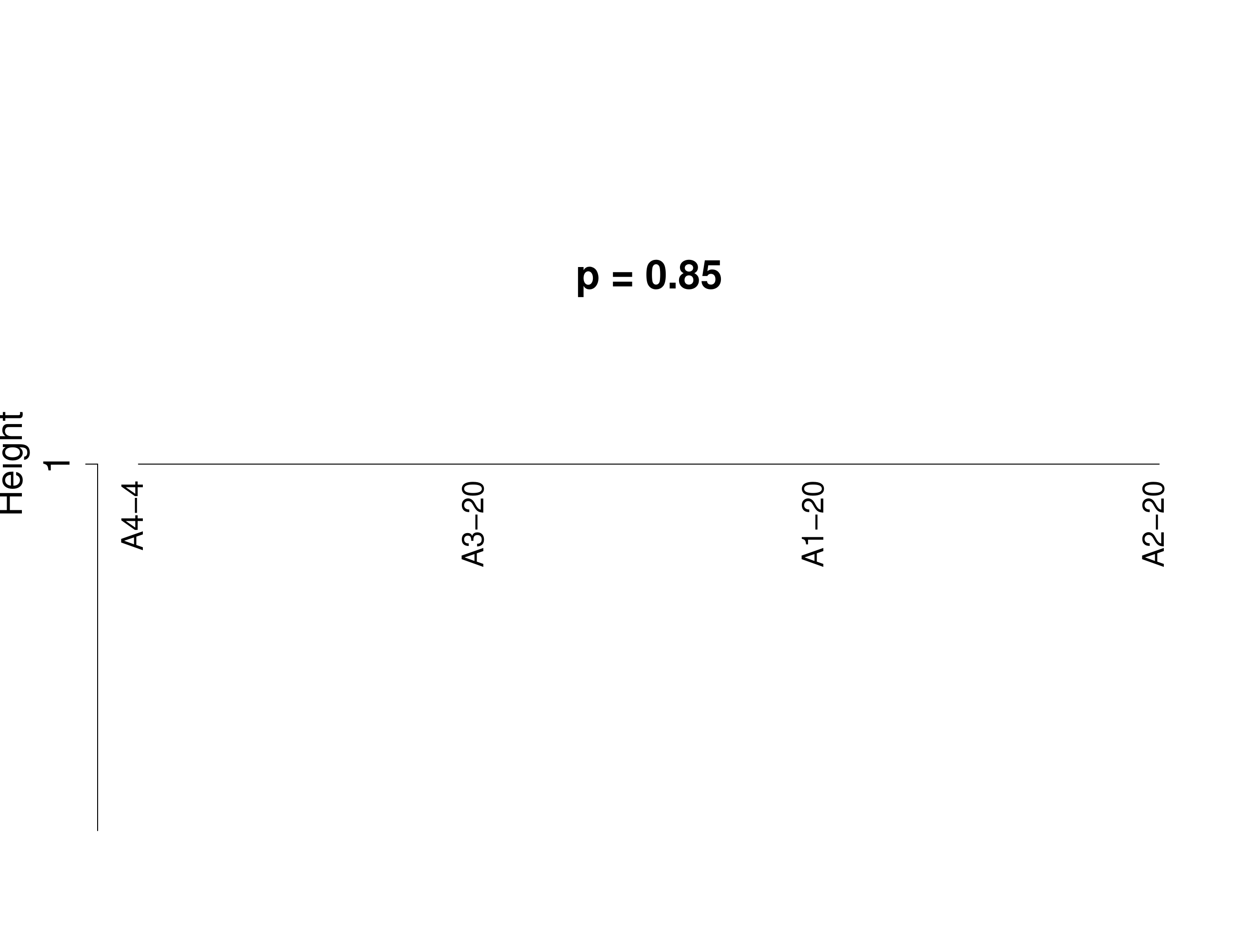}\\
{\small (c)}\\
\end{center}
\caption{Typical samples of attractor dendrograms.}
\label{fig:dendrograms}
\end{figure}
A similar picture emerges from the dendrograms, which graphically capture the clusters emerging among attractors. In Figures~\ref{fig:dendrograms}(a), \ref{fig:dendrograms}(b) and \ref{fig:dendrograms}(c), typical cases of dendrograms for chaotic, critical and ordered BNs are respectively plotted.

\section{Conclusion and future work}
\label{sec:conclusion}

In this work, we have studied some relevant statistical features of the distances among attractors in RBNs. We observed that chaotic networks have a scattered attractors set, while ordered networks' attractors show a strong tendency to form a cluster; critical networks exhibit a pattern that is a mixture of the two previous cases. This contribution is a first step towards the study of attractor sets and landscapes in RBNs with the aim of testing whether this GRN model is suitable for reproducing features of real cells. A further step will be the comparison against data of expression levels of genes in real cells.

\end{document}